\documentclass[preprint,12pt,a4paper]{elsarticle}

\usepackage{amssymb}
\usepackage{amsmath}
\usepackage{hyperref}
\usepackage{graphicx}
\usepackage{booktabs}
\usepackage{siunitx}
\usepackage{subfig}
\DeclareSIUnit{\fps}{fps}

\journal{SoftwareX}

\begin{document}

\begin{frontmatter}

\title{ARCANE-PedSynth: Synthetic Multi-Pedestrian Datasets with Behavioural Crossing Annotations}

\author[label1]{Muhammad Naveed Riaz}
\author[label2]{Maciej Wielgosz\corref{cor1}}
\ead{wielgosz@agh.edu.pl}
\author[label1]{Antonio M.\ L\'{o}pez Pe\~{n}a}

\cortext[cor1]{Corresponding author.}

\address[label1]{Departament de Ci\`{e}ncies de la Computaci\'{o} and Computer Vision Center (CVC),
  Universitat Aut\`{o}noma de Barcelona (UAB), Bellaterra, Barcelona, Spain}
\address[label2]{Institute of Electronics, Faculty of Computer Science, Electronics and Telecommunications,
  AGH University of Krakow, Krak\'{o}w, Poland}

\begin{abstract}
We present ARCANE-PedSynth, an open-source CARLA-based software framework for generating synthetic multi-pedestrian datasets with dense behavioural annotations for pedestrian crossing prediction in autonomous driving. The framework overcomes CARLA's native 9\% crossing rate through a hybrid AI-manual pedestrian control architecture, enabling configurable target rates up to 75\%. A 12-state behavioural finite state machine with five character archetypes produces diverse crossing behaviours. The framework generates synchronised RGB, LiDAR, and DVS data with per-frame crossing labels, behavioural states, and estimated 2D pose keypoints. We demonstrate ARCANE-PedSynth through PedSynth++, an example dataset generated with the framework, comprising 533 multi-pedestrian clips across 12 weather conditions with RGB, LiDAR, and DVS streams. ARCANE-PedSynth is fully reproducible via CLI parameterisation and Docker containerisation.
\end{abstract}


\end{frontmatter}


\section*{Metadata}
\label{sec:metadata}

Table~\ref{codeMetadata} summarises the code metadata for the ARCANE-PedSynth software release.

\begin{table}[!h]
\begin{tabular}{|l|p{6.5cm}|p{6.5cm}|}
\hline
\textbf{Nr.} & \textbf{Code metadata description} & \textbf{Metadata} \\
\hline
C1 & Current code version & v2.0.0 (Git tag; includes submodules \texttt{pedestrians-scenarios} and \texttt{pedestrians-video-2-carla}; clone with \texttt{--recurse-submodules}) \\
\hline
C2 & Permanent link to code/repository used for this code version & \url{https://github.com/wielgosz-info/carla-pedestrians}; 


\\
\hline
C3 & Permanent link to Reproducible Capsule & -- \\
\hline
C4 & Legal Code License & MIT Licence (core framework). Third-party components (OpenPose, SMPL/VPoser) have separate licences; see repository README for details. \\
\hline
C5 & Code versioning system used & git \\
\hline
C6 & Software code languages, tools, and services used & Python 3.8, CARLA 0.9.13, PyTorch, Docker, Docker Compose \\
\hline
C7 & Compilation requirements, operating environments \& dependencies & Linux (Ubuntu 20.04), NVIDIA GPU with CUDA 11.x, Docker $\geq$ 20.10, CARLA 0.9.13 server \\
\hline
C8 & If available, link to developer documentation/manual & \url{https://github.com/wielgosz-info/carla-pedestrians/blob/main/README.md} \\
\hline
C9 & Support email for questions & wielgosz@agh.edu.pl \\
\hline
\end{tabular}
\caption{Code metadata}
\label{codeMetadata}
\end{table}


\section{Motivation and significance}
\label{sec:motivation}

Predicting whether a pedestrian will cross in front of an autonomous vehicle (AV) is a safety-critical perception task. The vulnerability is particularly acute in urban environments where complex vehicle-pedestrian interactions occur frequently. Accurate crossing/non-crossing (C/NC) prediction enables proactive braking and comfortable maneuver planning; errors directly translate to collision risk. Unlike static obstacles, pedestrian behaviour exhibits high variability influenced by demographics, environmental conditions, traffic patterns, and individual risk perception~\cite{rasouli2020pedestrian}.

Real-world pedestrian datasets such as JAAD~\cite{rasouli2017are,rasouli2017agreeing} and PIE~\cite{rasouli2019PIE} have been instrumental in advancing the field, providing behavioural annotations from dashboard cameras~\cite{kotseruba2021benchmark}. However, crossing events constitute fewer than 10\% of observed frames, and safety-critical behaviours such as jaywalking, sudden dashes, and mid-road pauses are severely underrepresented. Manual annotation is expensive and prone to inconsistencies in annotation criteria, particularly for subjective attributes such as crossing intention and attention state. Moreover, real-world recording cannot ethically stage hazardous edge cases. The nuScenes dataset~\cite{caesar2020nuscenes} provides multi-modal sensor data but lacks C/NC annotations; the TITAN dataset~\cite{malla2020titan} offers detailed pedestrian actions but is not designed for binary crossing prediction. Several pose-based approaches~\cite{cadena2019pedestrian,lorenzo2020rnn} have demonstrated that skeleton keypoints provide valuable cues for crossing intention, motivating the integration of 2D pose estimates into synthetic data pipelines.

Synthetic data generation via simulators such as CARLA~\cite{dosovitskiy17carla} offers a principled path forward, producing unlimited data with pixel-perfect ground truth, controllable class balance, and systematic coverage of rare events. Earlier synthetic datasets—Virtual KITTI~\cite{gaidon2016virtualworlds}, SYNTHIA~\cite{ros2016synthia}, and GTA-V-based collections~\cite{richter2016gta}—demonstrated the feasibility of training perception models on simulator output.

More recently, HABIT introduced a CARLA benchmark that retargets real human motion from motion capture and video into physically consistent pedestrian behaviours, highlighting the importance of realistic interactive pedestrian motion for evaluating autonomous-driving agents~\cite{Ramesh_2026_WACV}.

In the pedestrian domain, Virtual-Pedcross-4667~\cite{vpedcross} introduced CARLA-generated clips with C/NC labels and a 61\% crossing ratio, but pedestrians follow scripted paths without naturalistic pre-crossing behaviours (hesitation, traffic checking, retreat) and the pipeline lacks multi-modal sensing or parametric control. DVS-PedX~\cite{dvspedx2026} provides paired synthetic (CARLA) and real (JAAD-to-DVS conversion) event-camera sequences with binary crossing labels, advancing event-based pedestrian perception; however, it focuses on single or limited pedestrian scenarios without behavioural FSM modelling, LiDAR sensing, or controllable crossing-rate generation. Critically, CARLA's built-in AI pedestrian system routes walkers along sidewalks, yielding a baseline crossing rate of only 9\%---insufficient for training balanced C/NC models.

We previously developed the ARCANE framework and used it to generate PedSynth~\cite{riaz2023pedsynth}: 947 single-pedestrian, RGB-only clips (approximately 20\,s each at 30\,FPS, totalling 5.5 hours) with binary C/NC labels. PedSynth demonstrated that combining synthetic and real data improves C/NC prediction, particularly for rare crossing events, and can be used to develop consistent C/NC self-annotation procedures~\cite{RiazSSL}. However, its single-agent, stationary-vehicle setup does not capture the complexity of real driving, where multiple pedestrians interact with dynamic traffic simultaneously. To our knowledge, no existing open-source tool provides all of the following: (a)~controllable crossing rates, (b)~multi-pedestrian dynamic traffic, (c)~multi-modal sensing, (d)~dense behavioural annotations beyond binary labels, and (e)~a reproducible open-source generation pipeline. Table~\ref{tab:comparison} compares representative datasets and frameworks across modality, annotation, and scenario dimensions.

\begin{table*}[!t]
\centering
\caption{Comparison of pedestrian crossing datasets and frameworks. \checkmark\ = fully supported, -- = not supported. ``Action labels'' indicates action classes are provided but not binary crossing/non-crossing labels. ``Partial'' behavioural states denote coarse attributes such as crossing intention or attention rather than explicit per-frame finite-state-machine states. ``Estimated'' pose indicates 2D keypoints obtained from RGB frames using external pose estimators rather than simulator-native or motion-capture skeleton ground truth; ARCANE-PedSynth currently uses this post-processing route for consistency with prior C/NC pipelines.}
\label{tab:comparison}
\resizebox{\textwidth}{!}{%
\begin{tabular}{lccccccccc}
\toprule
\textbf{Dataset} & \textbf{Synthetic} & \textbf{C/NC Labels} & \textbf{Multi-Ped.} & \textbf{Moving Ego} & \textbf{RGB} & \textbf{LiDAR} & \textbf{DVS} & \textbf{Pose} & \textbf{Behav.\ States} \\
\midrule
JAAD~\cite{rasouli2017are}               & --         & \checkmark     & \checkmark & \checkmark & \checkmark & --         & --         & Estimated & Partial    \\
PIE~\cite{rasouli2019PIE}                & --         & \checkmark     & \checkmark & \checkmark & \checkmark & --         & --         & Estimated & Partial    \\
nuScenes~\cite{caesar2020nuscenes}       & --         & --             & \checkmark & \checkmark & \checkmark & \checkmark & --         & --        & --         \\
TITAN~\cite{malla2020titan}              & --         & Action labels  & \checkmark & \checkmark & \checkmark & --         & --         & --        & \checkmark \\
V-Pedcross-4667~\cite{vpedcross}         & \checkmark & \checkmark     & Limited    & --         & \checkmark & --         & --         & --        & --         \\
HABIT~\cite{Ramesh_2026_WACV}            & \checkmark & Action labels  & \checkmark & \checkmark & \checkmark & --         & --         & MoCap     & \checkmark \\
DVS-PedX~\cite{dvspedx2026}              & \checkmark & \checkmark     & Limited    & \checkmark & \checkmark & --         & \checkmark & --        & --         \\
PedSynth~\cite{riaz2023pedsynth}         & \checkmark & \checkmark     & --         & --         & \checkmark & --         & --         & Estimated & --         \\
\textbf{PedSynth++}                      & \checkmark & \checkmark     & \checkmark & \checkmark & \checkmark & \checkmark & \checkmark & Estimated & \checkmark \\
\bottomrule
\end{tabular}%
}
\end{table*}

ARCANE-PedSynth addresses all of these gaps. The software has already been used to generate the PedSynth++ dataset (533 clips across 12 weather conditions with three sensor modalities), on which the PedGT model~\cite{pedgt} achieves F1=89 on the held-out Town05 test split after training only on synthetic PedSynth++ clips from Town01 and Town02. The framework is designed for reuse: researchers can generate arbitrarily many custom dataset variants by changing CLI parameters without modifying the source code.


\begin{figure}[!htbp]
\centering
\includegraphics[width=\columnwidth]{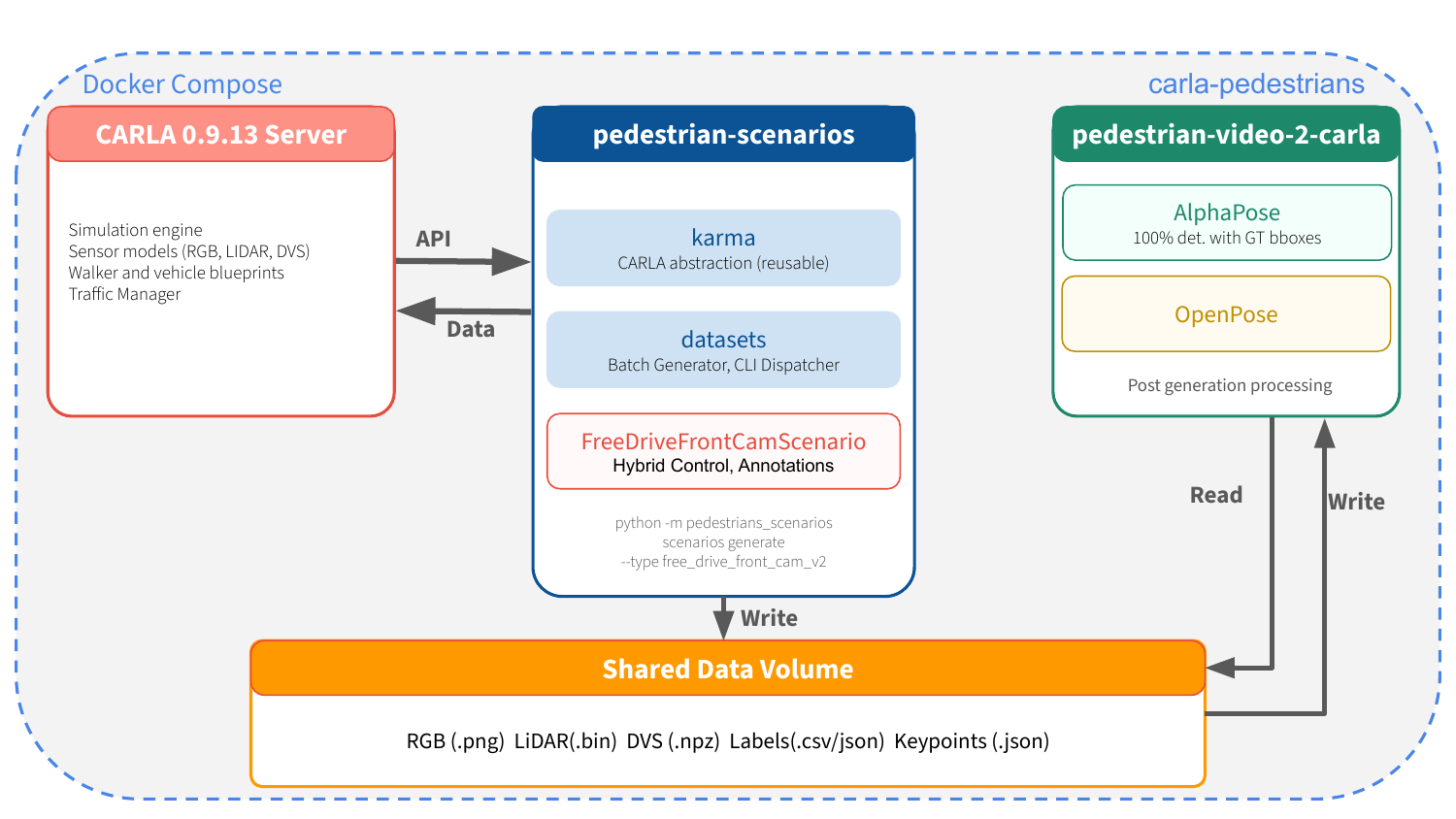}
\caption{Software architecture of ARCANE-PedSynth. Three Git 
repositories are orchestrated via Docker Compose (\texttt{carla-pedestrians}): the unmodified 
CARLA server, the \texttt{pedestrians-scenarios} generation 
container, and the \texttt{pedestrians-video-2-carla} pose 
extraction container. Pose keypoints are extracted from rendered RGB frames, rather than taken directly from CARLA's internal skeleton, to keep the annotation pipeline consistent with real-video processing and reduce a potential synthetic-to-real gap; see Sec.~\ref{sssec:pose_extraction}. Containers communicate through Docker 
networking and share generated data via mounted volumes.}
\label{fig:architecture}
\end{figure}

\section{Software description}
\label{sec:software}

\subsection{Software architecture}
\label{ssec:architecture}

ARCANE-PedSynth is organized into three Git repositories orchestrated via Docker Compose (Figure~\ref{fig:architecture}). Contrary to the common approach of adding custom code to the CARLA base image—which generates dependency conflicts—we keep the CARLA server in its original container and connect purpose-built containers through Docker's private networking and shared volumes:

\begin{itemize}
\item \textbf{\texttt{carla-pedestrians}} (main repo): Docker Compose orchestration of all containers, including the CARLA 0.9.13 server, scenario generation, and pose extraction services. The \texttt{docker-compose.yml} defines inter-container networking, GPU passthrough, and shared data volumes.
\item \textbf{\texttt{pedestrians-scenarios}}: Dataset generation logic containing two modules. The \texttt{karma} module provides low-level CARLA abstractions (coordinate transforms, actor lifecycle management, sensor configuration) reusable across projects. The \texttt{datasets} module provides the actual generation flow via \texttt{BatchGenerator} classes, with customisation done at the scenario level. New scenario types are implemented as Python classes registered through the CLI dispatcher.
\item \textbf{\texttt{pedestrians-video-2-carla}}: ML model training and inference flows using PyTorch and PyTorch Lightning, including post-generation pose extraction via AlphaPose~\cite{fang2017alphapose} and OpenPose~\cite{cao2021openpose}. This container requires the CARLA server only when rendering outputs back into the simulator; for pose extraction, it operates independently on pre-rendered RGB frames.
\end{itemize}

\begin{figure}[!htbp]
    \centering
    \includegraphics[width=1.0\textwidth]{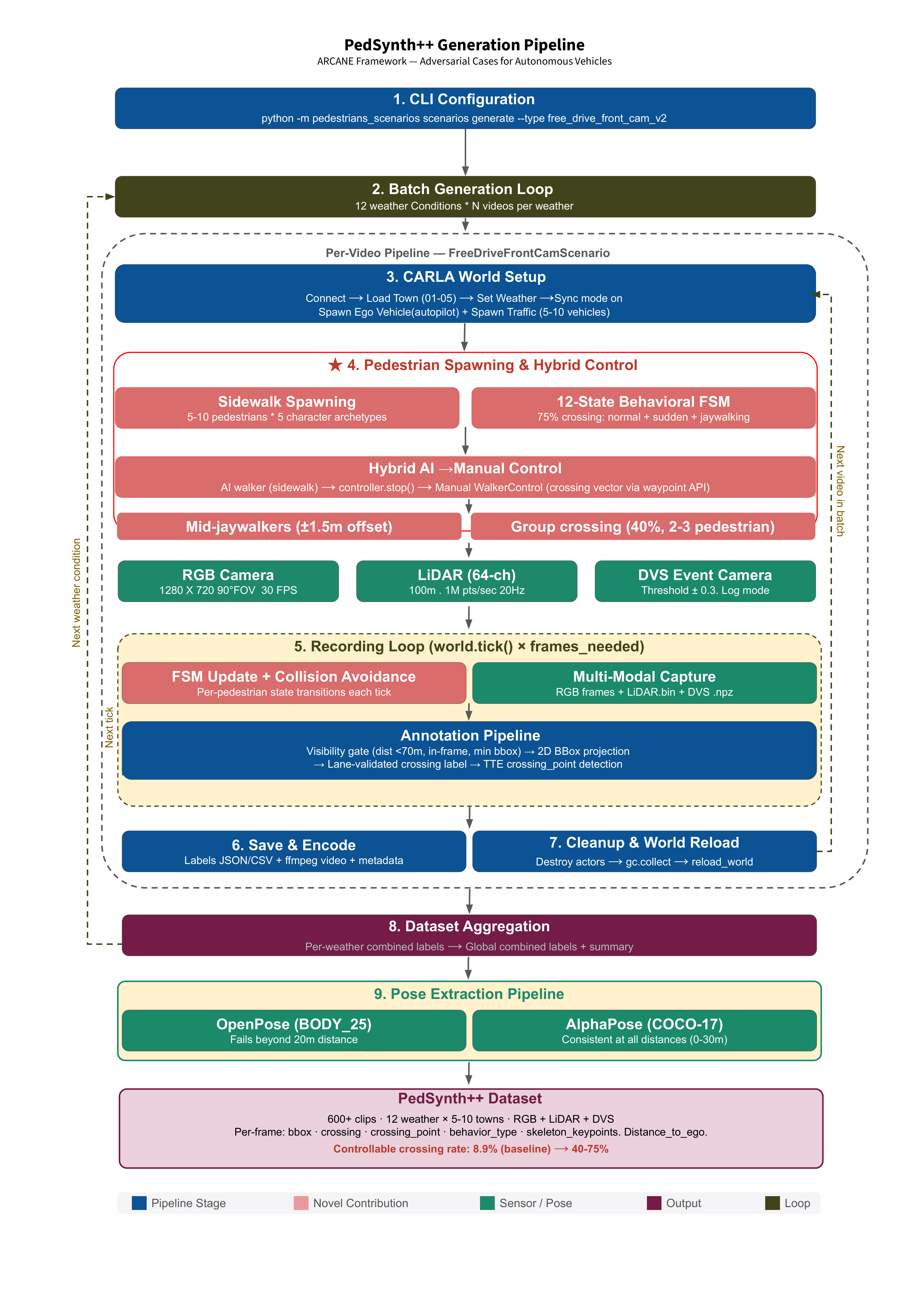}
    \caption{PedSynth++ generation pipeline within the ARCANE PedSynth. Nine stages from CLI configuration to final dataset output. Red-highlighted Stage~4 indicates the core novelty: 
    hybrid AI-manual pedestrian control. Dashed arrows indicate iteration loops across weather conditions, videos, and simulation ticks.}
    \label{fig:scenario_overview}
\end{figure}

The core generation is invoked through the standard CLI:
\begin{verbatim}
python -m pedestrians_scenarios scenarios generate \
  --type free_drive_front_cam_v2 [options]
\end{verbatim}

Key CLI parameters include: \texttt{--dataset\_\allowbreak mode} (batch across weather conditions), \texttt{--videos\_\allowbreak per\_\allowbreak weather}, \texttt{--crossing\_\allowbreak ratio}, \texttt{--sudden\_\allowbreak crossing\_\allowbreak ratio}, \texttt{--jaywalking\_\allowbreak ratio}, \texttt{--enable\_\allowbreak lidar}, \texttt{--enable\_\allowbreak dvs}, and \texttt{--enable\_\allowbreak emergency}.

\subsection{Software functionalities}
\label{ssec:functionalities}

\subsubsection{Hybrid pedestrian control}
The central technical contribution overcomes CARLA's 9\% crossing baseline through a dual-mode architecture. This low rate is a fundamental limitation of CARLA's navigation mesh, which treats road surfaces as obstacles for pedestrian pathfinding. Therefore, simply spawning pedestrians and relying on AI navigation produces almost exclusively non-crossing behaviour.

In \emph{AI mode}, each pedestrian is assigned a CARLA AI walker controller via \texttt{controller.start()} and directed to walk along sidewalks using \texttt{go\_\allowbreak to\_\allowbreak location()}. The AI controller handles path planning, obstacle avoidance, and speed regulation, producing a natural walking animation. In \emph{manual mode}, the AI controller is stopped via \texttt{controller.stop()} and the pedestrian is controlled directly through \texttt{carla.WalkerControl} commands specifying direction vectors and speeds. This enables precise control over crossing trajectory, speed profile, mid-crossing pauses, and retreat behavior. The system computes a perpendicular crossing vector by querying CARLA's waypoint API for the nearest road waypoint and computing the direction perpendicular to the road's forward vector.

A four-layer crossing-rate engineering strategy (Table~\ref{tab:crossing_rate}) progressively raises the effective rate from CARLA's 9\% baseline to 40--75\%. Each layer is independently configurable via the CLI, enabling researchers to fine-tune the crossing distribution for specific experimental needs.

\begin{table}[t]
\centering
\caption{Four-layer crossing rate engineering strategy.}
\label{tab:crossing_rate}
\renewcommand{\arraystretch}{1.2}
\resizebox{\columnwidth}{!}{%
\begin{tabular}{cllll}
\toprule
\textbf{Layer} & \textbf{Mechanism} & \textbf{Parameter} & \textbf{Value} & \textbf{Effect} \\
\midrule
1 & Crosser allocation at spawn   & \texttt{crosser\_\allowbreak ratio}       & 90\%      & Sidewalk positions 10--55\,m ahead \\
2 & Behavioural type assignment    & \texttt{crossing\_\allowbreak behaviors}  & 75\%      & 50\% normal / 25\% sudden / 25\% jaywalk \\
3 & Mid-road jaywalker injection   & \texttt{lateral\_\allowbreak offset}      & $\pm$1.5\,m & Spawn directly in driving lanes \\
4 & Group crossing formation       & \texttt{group\_\allowbreak probability}   & 40\%      & Synchronised groups of 2--3 \\
\bottomrule
\end{tabular}%
}
\end{table}

\subsubsection{Behavioural finite state machine}
Each pedestrian is driven by a 12-state FSM (Table~\ref{tab:fsm_states}) whose transitions depend on the local traffic context and a fixed \emph{character archetype} (Table~\ref{tab:archetypes}). Five archetypes \texttt{business\_\allowbreak person}, \texttt{casual\_\allowbreak person}, \texttt{elderly\_\allowbreak person}, \texttt{young\_\allowbreak person}, and \texttt{parent\_\allowbreak with\_\allowbreak child}---span the demographic spread from cautious to risk-taking pedestrians, with proportions and behavioral parameters informed by observational studies of pedestrian behavior~\cite{rasouli2020pedestrian,ishaque2008behavioural,papadimitriou2009meta}. The archetype fixes each agent's base walking speed, hesitation duration, attention level, and intrinsic crossing propensity; the FSM then decides, at every simulation tick, when to look around, check traffic, hesitate, cross, pause mid-road, retreat, or finish the crossing.

\paragraph{Gap acceptance via time-to-collision}
The decision to step onto the road is governed by a gap-acceptance test based on time-to-collision (TTC). For each pedestrian position~$\mathbf{p}$ and each vehicle within sensor range (position~$\mathbf{q}$, velocity~$\mathbf{v}_{\text{veh}}$), we compute the unit direction $\hat{\mathbf{d}} = (\mathbf{p}-\mathbf{q})/\|\mathbf{p}-\mathbf{q}\|$ and the projected closing speed $v_{\text{close}} = \mathbf{v}_{\text{veh}} \cdot \hat{\mathbf{d}}$. TTC is defined only when the vehicle is actually closing the gap:
\[
\text{TTC} =
\begin{cases}
\|\mathbf{p}-\mathbf{q}\| \;/\; v_{\text{close}}, & v_{\text{close}} > 0, \\
\infty, & \text{otherwise.}
\end{cases}
\]
The check is applied to \emph{all} traffic vehicles in range, not only the ego-vehicle. A crossing is initiated when the minimum TTC across vehicles exceeds the time required to traverse the road plus a safety margin,
\[
\min_{\text{veh}} \text{TTC} \;>\; \tau_{\text{ttc}} + \Delta_{\text{safety}} + w_{\text{road}}/s_{\text{cross}},
\]
where $\tau_{\text{ttc}}\!\in\![3,6]\,\text{s}$ is an archetype-specific gap-acceptance threshold, $\Delta_{\text{safety}}\!\in\![0.5,1.0]\,\text{s}$ is a safety buffer, $w_{\text{road}}$ is the local road width, and $s_{\text{cross}}$ is the pedestrian's crossing speed.

\paragraph{In-crossing safety and patience}
Once a pedestrian has entered the road, vehicle proximity is re-evaluated at three lookahead points along the crossing vector (0.5, 1.5, and 2.5\,m ahead) and at a predicted position 0.5\,s in the future; if any vehicle falls within a 3\,m safety radius of these points, the agent halts and may enter \texttt{PAUSING\_\allowbreak MID\_\allowbreak CROSS} or, with a 15\,\% probability while still in the first 40\,\% of the crossing path, \texttt{RETREAT}—producing the mid-road reversals and near-misses that are particularly valuable for safety-critical training. Patience is bounded by an archetype-dependent waiting-time cap (6--18\,s); when it expires, the pedestrian either abandons the attempt or forces a \texttt{SUDDEN\_\allowbreak CROSSING}. In practice, a potential crosser whose archetype check has succeeded commits to entering the road as soon as the ego-vehicle is within the 10--40\,m decision window, simulating the typical urban interaction geometry between approaching vehicles and pedestrians waiting at the curb.

\paragraph{Group crossings and situational speed}
To reproduce the social structure of urban pedestrian flow, potential crossers are paired into groups of two or three, with a 40\,\% probability that group members share a crossing-group identifier and a leader, which synchronizes their state transitions and yields coherent group crossings rather than independent walkers. Orthogonally, every pedestrian selects one of four situational speed profiles that can switch during a clip\emph{—cautious} ($0.8\times$, poor visibility or elderly agents), \emph{normal} ($1.0\times$), \emph{rushed} ($1.3\times$, when closing a tight gap), and \emph{distracted} ($0.9\times$, with occasional pauses to simulate phone use)—so that base speed (archetype) and instantaneous speed (situation) are decoupled, yielding the speed variability observed in real crossing footage without per-agent hand-tuning.

\begin{table}[t]
\centering
\caption{Behavioural state machine: 12 states.}
\label{tab:fsm_states}
\renewcommand{\arraystretch}{1.1}
\resizebox{\columnwidth}{!}{%
\begin{tabular}{lll}
\toprule
\textbf{State} & \textbf{Description} & \textbf{Trigger} \\
\midrule
\texttt{WALKING\_\allowbreak SIDEWALK}   & AI-controlled navigation            & Timer / crossing trigger \\
\texttt{LOOKING\_\allowbreak AROUND}     & Visual scanning                     & Attention threshold \\
\texttt{CHECKING\_\allowbreak TRAFFIC}   & Assessing vehicle proximity         & TTC calculation \\
\texttt{HESITATING}          & Pausing at road edge                & Confidence threshold \\
\texttt{CROSSING\_\allowbreak ROAD}      & Steady crossing at base speed       & Road edge reached, gap accepted \\
\texttt{SUDDEN\_\allowbreak CROSSING}    & Abrupt entry without checking       & Random trigger \\
\texttt{JAYWALKING}          & Mid-block crossing                  & Position-based \\
\texttt{RUNNING\_\allowbreak ACROSS}     & High-speed crossing                 & Vehicle proximity \\
\texttt{PAUSING\_\allowbreak MID\_\allowbreak CROSS} & Stopping mid-road                   & Distraction / fear \\
\texttt{DISTRACTED\_\allowbreak BEHAVIOR}& Phone use / inattention             & Archetype probability \\
\texttt{FINISHED\_\allowbreak CROSSING}  & Reached opposite sidewalk           & Lane exit detection \\
\texttt{RETREAT}             & Mid-crossing reversal               & Fast approaching vehicle \\
\bottomrule
\end{tabular}%
}
\end{table}

\begin{table}[t]
\centering
\caption{Character archetypes and behavioural parameters.}
\label{tab:archetypes}
\renewcommand{\arraystretch}{1.1}
\resizebox{\columnwidth}{!}{%
\begin{tabular}{lccccc}
\toprule
\textbf{Archetype} & \textbf{Proportion} & \textbf{Speed (m/s)} & \textbf{Hesitation} & \textbf{Attention} & \textbf{Cross.\ Prob.} \\
\midrule
\texttt{business\_\allowbreak person}    & 30\% & 1.4--1.8 & 0.3--0.8\,s & 0.5 & 0.85 \\
\texttt{casual\_\allowbreak person}      & 25\% & 1.0--1.4 & 0.5--1.5\,s & 0.7 & 0.70 \\
\texttt{elderly\_\allowbreak person}     & 10\% & 0.6--1.0 & 1.0--2.0\,s & 0.9 & 0.50 \\
\texttt{young\_\allowbreak person}       & 20\% & 1.2--2.0 & 0.2--0.6\,s & 0.4 & 0.90 \\
\texttt{parent\_\allowbreak with\_\allowbreak child} & 15\% & 0.8--1.2 & 0.8--1.8\,s & 0.8 & 0.60 \\
\bottomrule
\end{tabular}%
}
\end{table}

\subsubsection{Multi-modal sensing and annotation}
The framework captures three synchronized sensor modalities mounted on the ego-vehicle: an RGB camera (1280$\times$720, 90$^\circ$~FOV, 30\,FPS) at the driver's eye level, a 64-channel LiDAR (100\,m range, 1M~points/s, 20\,Hz, 360$^\circ$~horizontal) on the roof mount, and a DVS event camera co-located with the RGB camera. The LiDAR provides 3D point clouds whose geometric measurements remain reliable regardless of lighting conditions, complementing camera-based perception under adverse weather. The DVS sensor captures asynchronous brightness changes with microsecond resolution, yielding extremely low latency and high dynamic range—properties particularly valuable for detecting fast-moving pedestrians at dusk or under flickering artificial illumination.

A four-stage visibility gate determines which pedestrians are annotated: (1) distance to ego $<$ 70\,m, (2) positive dot product with camera forward vector, (3) projected bounding box $\geq$ 15$\times$30\,px (8$\times$15\,px beyond 50\,m), and (4) bounding box within image bounds. Annotations generated automatically during the synchronous recording loop include: lane-validated crossing labels, behavioral state, bounding boxes, distance to ego, time-to-event markers following JAAD/PIE conventions, character archetype, and 17-keypoint COCO-format 2D pose estimates (obtained via AlphaPose post-processing; see below).

\paragraph{Crossing label semantics}
A pedestrian is labeled as \emph{crossing} (C=1) in a given frame when two conditions are simultaneously satisfied: (i) the FSM is in a crossing-related state (\texttt{CROSSING\_\allowbreak ROAD}, \texttt{SUDDEN\_\allowbreak CROSSING}, \texttt{JAYWALKING}, \texttt{RUNNING\_\allowbreak ACROSS}, or \texttt{PAUSING\_\allowbreak MID\_\allowbreak CROSS}), \emph{and} (ii) the pedestrian's physical position falls on a \texttt{carla.LaneType.Driving} lane (with a short temporal smoothing window that keeps the label stable across brief lane-classification glitches on shoulders and parking strips). This dual condition avoids false positives from FSM state alone (e.g., a pedestrian that has decided to cross but has not yet stepped onto the road). Sequence-level labels follow the JAAD/PIE convention: a pedestrian is classified as a \emph{crosser} if it is labeled C=1 in any frame of its visible sequence. A \emph{time-to-event} (TTE) marker records the number of frames remaining until the first C=1 frame, enabling models to predict crossing \emph{intention} before the physical crossing begins.

\subsubsection{Post-generation pose extraction}
\label{sssec:pose_extraction}
While CARLA provides internal skeleton data (26 joints per pedestrian model), these are 3D model-space joints that require non-trivial projection and do not follow standard 2D pose formats. To obtain publication-quality COCO-17 format keypoints, the framework performs dedicated 2D pose estimation on the rendered RGB frames as a post-processing step. AlphaPose~\cite{fang2017alphapose} and OpenPose~\cite{cao2021openpose} are both integrated into the Docker infrastructure for automated batch processing. When supplied with simulator-generated ground-truth bounding boxes, AlphaPose produced a pose estimate for every annotated visible pedestrian in the processed dataset under generation and is the recommended extractor. OpenPose exhibits distance-dependent detection failures, particularly for pedestrians beyond 40\,m where projected bounding boxes become small. We emphasize that these are \emph{estimated} 2D keypoints, not ground-truth projections of CARLA's internal skeleton. However, the use of noise-free simulator bounding boxes provides an upper-bound detection setting: in real-world datasets, JAAD has 85.74\% missing joints and PIE has 93.28\% missing keypoints~\cite{kotseruba2021benchmark}.

\subsubsection{Memory-safe batch generation}
Generating hundreds of video clips requires careful resource management. CARLA accumulates internal state (actor references, physics state, rendering resources) that cannot be fully released through its API. After each video, the framework implements a six-phase cleanup protocol: (1)~Traffic Manager shutdown, (2)~sensor stop and destruction, (3)~AI controller termination, (4)~individual actor destruction with micro-sleeps between destructions to prevent race conditions, (5)~reference clearing, and (6)~Python garbage collection and GPU cache clearing. The world is then fully reloaded via \texttt{client.reload\_\allowbreak world()} with a 10-second stabilization period to prevent out-of-memory crashes during extended batch generation.


\section{Illustrative examples}

\label{sec:examples}
\begin{table}[t]
\centering
\caption{PedSynth++ dataset statistics.}
\label{tab:dataset_stats}
\renewcommand{\arraystretch}{1.2}
\begin{tabular}{lr}
\toprule
\textbf{Statistic} & \textbf{Value} \\
\midrule
Total video clips                    & 533 \\
Total frames                         & 177,231 \\
Avg.\ frames per clip                & 332 \\
Avg.\ clip duration                  & $\sim$11\,s at 30\,FPS \\
\midrule
Total unique pedestrians              & 3,426 \\
\quad Crossing (C)                    & 2,143 (62.5\%) \\
\quad Non-crossing (NC)               & 1,283 (37.5\%) \\
C/NC ratio                           & 1.67:1 \\
\midrule
Total annotated samples (ped-frames)  & 1,631,312 \\
Avg.\ samples per clip                & 3,061 \\
\midrule
Weather conditions                    & 12 \\
CARLA towns                          & 4 (Town01, 02, 03, 05) \\
Sensor modalities                     & 3 (RGB, LiDAR, DVS) \\
\bottomrule
\end{tabular}
\end{table}

\begin{figure*}[t]
\centering
\includegraphics[width=\textwidth]{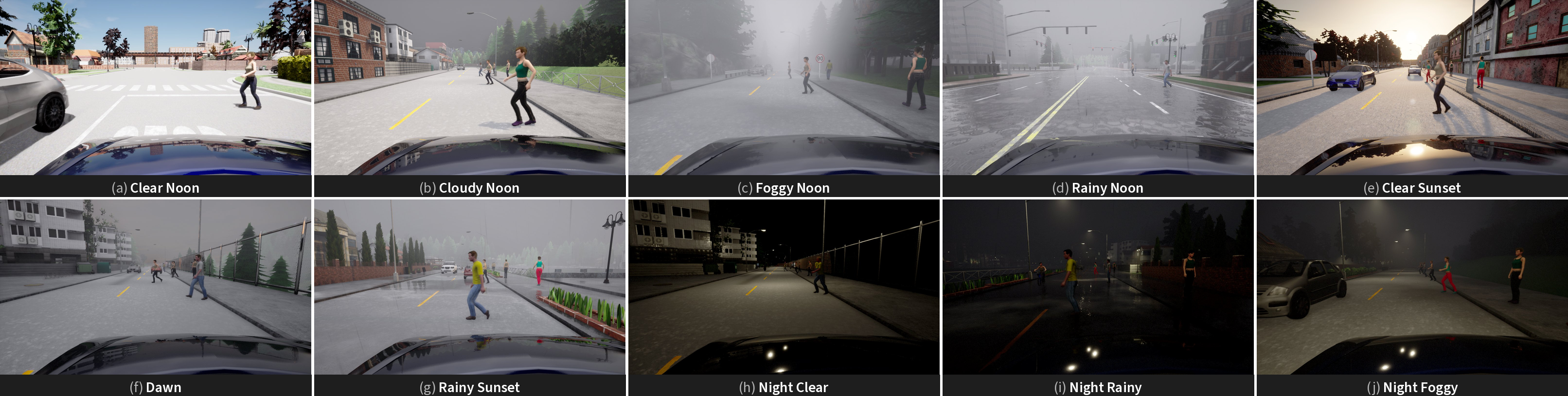}
\caption{Sample RGB frames from PedSynth++ across 10 weather 
conditions. Top row: daytime conditions from optimal visibility 
to low-angle lighting. Bottom row: transitional and 
nighttime conditions with increasing perceptual difficulty.}
\label{fig:weather_samples}
\end{figure*}

\textbf{Example 1: Generating the PedSynth++ dataset.}
Using ARCANE-PedSynth, we generated PedSynth++: 533 video clips across 12 weather conditions (Table~\ref{tab:weather_conditions}) and four CARLA towns (Town01, Town02, Town03, and Town05), providing diverse urban topologies including residential streets, commercial districts, multi-lane roads, and complex intersections. Figure~\ref{fig:weather_samples} shows representative RGB frames illustrating the visual diversity across weather and lighting conditions. Table~\ref{tab:dataset_stats} summarises the complete dataset statistics.

Each clip runs for approximately 10--15\,s at 30\, FPS (300--450 frames) and contains 5--10 pedestrians and 5--10 traffic vehicles with varied driving profiles (20\% aggressive, 60\% normal, 20\% cautious) alongside a moving ego-vehicle managed by CARLA's Traffic Manager. The total dataset comprises over 177,000 annotated frames with synchronised RGB, LiDAR, and DVS data. The controllable crossing rate of 40--75\% is the dataset's most distinctive characteristic---researchers can generate balanced 50/50 splits for standard training, imbalanced distributions matching real-world conditions for evaluation, or crossing-heavy distributions for focused rare-event training.

\begin{table}[t]
\centering
\caption{Twelve weather configurations spanning three difficulty levels.}
\label{tab:weather_conditions}
\resizebox{\columnwidth}{!}{%
\begin{tabular}{@{}lccc@{}}
\toprule
\textbf{Weather Type} & \textbf{Sun Alt.\ ($^\circ$)} & \textbf{Key Conditions} & \textbf{Difficulty} \\
\midrule
Clear Noon    & 70  & Optimal baseline              & Easy \\
Cloudy Noon   & 60  & Diffuse lighting              & Easy \\
Wet Noon      & 65  & Post-rain surface effects      & Easy \\
Soft Rain     & 60  & Intermediate precipitation     & Moderate \\
Foggy Noon    & 45  & Visibility limitation          & Moderate \\
Clear Sunset  & 10  & Extreme shadows                & Moderate \\
Night Clear   & $-$40 & Artificial lighting only      & Moderate \\
Dawn          & 5   & Transitional lighting          & Moderate \\
Heavy Rain    & 50  & Severe precipitation           & Hard \\
Rainy Sunset  & 12  & Poor light + precipitation    & Hard \\
Night Rainy   & $-$35 & Minimal light + rain + fog   & Hard \\
Night Foggy   & $-$35 & Extreme darkness + fog        & Hard \\
\bottomrule
\end{tabular}%
}
\end{table}

\textbf{Example 2: Training a state-of-the-art crossing prediction model.}
We trained PedGT~\cite{pedgt}, a lightweight graph-transformer architecture combining graph neural networks with a transformer encoder that captures spatio-temporal dependencies from pedestrian 2D pose keypoints and bounding boxes. Following established JAAD/PIE evaluation protocols~\cite{kotseruba2021benchmark}, we tested observation windows of 8, 16, and 30 input frames. The data was split by clip (not by frame or pedestrian) to prevent information leakage: clips from Town01 and Town02 form the training set, Town03 is used for validation, and Town05 for testing, ensuring that the model is evaluated on unseen urban layouts. The class distribution in the test set is approximately 60\%/40\% C/NC, reflecting the generator's target ratio.

The 16-frame model achieves the best F1=89 (Table~\ref{tab:input_frames}), with shorter and longer windows yielding higher recall but lower precision (over-prediction of crossing events). This result is consistent with findings on real-world datasets, where moderate temporal windows outperform both very short sequences (insufficient context) and very long ones (diluted signal). Notably, no data normalisation was required---the geometrically consistent annotations from ARCANE-PedSynth preserve absolute spatial relationships between joints (e.g., proximity to road edge, relative height changes during stepping off a curb) that aid crossing prediction. Aggressive normalisation schemes that discard positional information actually degrade performance.

These results validate the internal consistency and learnability of the generated annotations. They do not by themselves prove sim-to-real generalisation, which requires domain adaptation techniques and is the subject of ongoing work~\cite{RiazSSL}.

\begin{table}[t]
\centering
\caption{PedGT C/NC prediction on PedSynth++ data (no normalisation).}
\label{tab:input_frames}
\renewcommand{\arraystretch}{1.2}
\resizebox{\columnwidth}{!}{%
\begin{tabular}{cccccc}
\toprule
\textbf{Input Frames} & \textbf{Accuracy} & \textbf{Precision} & \textbf{Recall} & \textbf{F1} & \textbf{AUC} \\
\midrule
8  & 79.95 & 80 & 92 & 86 & 81 \\
16 & \textbf{85.45} & \textbf{87} & 91 & \textbf{89} & \textbf{89} \\
30 & 83.64 & 84 & 92 & 88 & 86 \\
\bottomrule
\end{tabular}%
}
\end{table}

\textbf{Example 3: Custom dataset generation.}
Researchers can generate variants without code changes. The following minimal command produces a small test dataset of two clips in a single town and weather condition:
\begin{verbatim}
python -m pedestrians_scenarios scenarios generate \
  --type free_drive_front_cam_v2 \
  --outputs_dir /outputs/test \
  --towns Town01 \
  --weather_conditions clear_noon \
  --videos_per_weather 2 \
  --crossing_ratio 0.6 \
  --enable_lidar --enable_dvs \
  --port 2000 --tm_port 8000 --host server
\end{verbatim}
For full-scale generation, a batch across all weather conditions is invoked with \texttt{--dataset\_\allowbreak mode}:
\begin{verbatim}
python -m pedestrians_scenarios scenarios generate \
  --type free_drive_front_cam_v2 \
  --outputs_dir /outputs/pedsynth_pp \
  --dataset_mode \
  --videos_per_weather 50 \
  --crossing_ratio 0.6 \
  --enable_lidar --enable_dvs \
  --port 2000 --tm_port 8000 --host server
\end{verbatim}
This command produces 600 clips (50~per weather $\times$ 12 conditions) with a 60\% target crossing rate and all three sensor modalities enabled.


\section{Impact}
\label{sec:impact}

ARCANE-PedSynth enables several research directions that were previously difficult or impossible:

\textbf{Controllable class balance for rare-event training.}
The four-layer crossing rate engineering provides, to our knowledge, the first open-source CARLA mechanism for configurable target crossing rates. The current implementation supports target rates between 40\% and 75\%. . This directly addresses the class imbalance problem that limits real-world datasets such as JAAD and PIE (fewer than 10\% crossing frames). Researchers can generate balanced training sets, imbalanced evaluation sets matching real-world distributions, or crossing-heavy sets for focused rare-event training.

\textbf{Behavioural diversity beyond binary labels.}
The 12-state FSM with five character archetypes produces behaviours absent from most existing datasets: hesitation, traffic checking, mid-road pausing, retreat, and group crossings. These fine-grained behavioural annotations support research on intention prediction, risk assessment, and behavioural modelling beyond simple C/NC classification. The archetype-specific parameters---informed by observational studies of real pedestrian behaviour~\cite{rasouli2020pedestrian,ishaque2008behavioural,papadimitriou2009meta}---span the population distribution from cautious elderly pedestrians to risk-taking young adults.

\textbf{Multi-modal fusion research.}
Synchronised RGB, LiDAR, and DVS outputs from the same scenes enable sensor fusion research for pedestrian safety. LiDAR provides depth-invariant 3D information; DVS event cameras capture motion with microsecond resolution under challenging lighting — both complement RGB-based perception. The 12 weather configurations systematically vary conditions from optimal (Clear Noon) to extreme (Night Foggy), enabling rigorous evaluation of weather-dependent performance degradation.

\textbf{Adversarial and domain adaptation training.}
The FSM's rare-event states (\texttt{SUDDEN\_\allowbreak CROSSING}, \texttt{RETREAT}, \texttt{PAUSING\_\allowbreak MID\_\allowbreak CROSS}) generate corner cases that challenge AV robustness. The earlier PedSynth dataset has already demonstrated that combined synthetic-real training outperforms real-only baselines~\cite{riaz2023pedsynth}. The dataset has also been employed to reduce manual annotation 
requirements for pedestrian intention prediction, enabling 
semi-supervised labelling workflows where synthetic annotations 
serve as pseudo-labels for real-world data~\cite{RiazSSL}.   PedSynth++'s multi-modal nature may further support domain adaptation research by providing complementary signal types with different domain gap characteristics.

\textbf{In-domain validation.}
The PedSynth++ dataset generated by this framework has been used to train PedGT~\cite{pedgt} to F1=89 on the held-out Town05 test split, confirming that the behavioural diversity and balanced crossing rate translate to effective in-domain C/NC prediction. The framework supports pose-based approaches~\cite{cadena2019pedestrian,lorenzo2020rnn} through integrated AlphaPose 2D pose extraction: when supplied with simulator-generated ground-truth bounding boxes, AlphaPose produced a pose estimate for every annotated visible pedestrian---substantially better than detection rates on real-world datasets, where JAAD has 85.74\% missing joints in pose data~\cite{kotseruba2021benchmark}.

\textbf{Reproducibility and extensibility.}
Docker containerisation ensures consistent CUDA versions, CARLA API compatibility, and Python dependency management across machines. The separation between the CARLA server container and the generation container avoids the dependency conflicts common in monolithic setups. New scenario types, behavioural models, or sensor configurations can be added by subclassing existing Python classes without modifying the core framework. The core ARCANE-PedSynth code is released under the MIT Licence. However, optional third-party components used for pose extraction or body-model processing---including OpenPose, SMPL/SMPL-X, and VPoser-related assets---are governed by their own licences and may restrict commercial use. Users should consult the repository README and individual component licences before commercial deployment.


\section{Limitations}
\label{sec:limitations}

The generated data remains synthetic and inherits the behavioural limitations of the CARLA simulator. Although the FSM increases crossing diversity, it does not fully model the richness of real pedestrian interaction, including pedestrian-pedestrian negotiation, gaze-based communication with drivers, social group dynamics, or culturally and contextually dependent crossing norms. The five character archetypes, while informed by observational literature~\cite{rasouli2020pedestrian,ishaque2008behavioural,papadimitriou2009meta}, are simplified behavioural profiles; real pedestrian populations exhibit a continuous spectrum of risk tolerance, mobility constraints, attention, and situational awareness. Interactions with vehicles are governed by rule-based gap acceptance and traffic-manager behaviour rather than learned social reasoning, so rare negotiation patterns such as ambiguous yielding, hesitation caused by driver intent, or collective crossing decisions remain only partially represented. Finally, the PedGT results reported on PedSynth++ (Sec.~\ref{sec:examples}) demonstrate in-domain learnability of the generated annotations but do not alone establish real-world generalisation; sim-to-real transfer remains an active research direction~\cite{RiazSSL}.


\section{Conclusions}
\label{sec:conclusions}

We have presented ARCANE-PedSynth, an open-source CARLA-based framework for generating synthetic multi-pedestrian datasets with dense behavioural annotations for pedestrian crossing prediction in autonomous driving. Its main contribution is a reproducible software pipeline for controllable multi-pedestrian crossing generation with behavioural FSM states, synchronised RGB/LiDAR/DVS sensing, and dense per-frame annotations, enabling researchers to generate custom datasets with controlled crossing-rate distributions and rare safety-critical behaviours. The software's hybrid AI-manual pedestrian control architecture overcomes a fundamental limitation of CARLA's navigation mesh---its native 9\% crossing rate---enabling configurable target rates up to 75\% through a principled four-layer crossing-rate engineering strategy. A 12-state behavioural FSM driven by TTC-based gap-acceptance modelling, with five character archetypes spanning realistic demographic distributions, produces diverse crossing behaviours including hesitation, traffic checking, mid-road pausing, retreat, and group crossings. Synchronised RGB, LiDAR, and DVS sensing with comprehensive automatic per-frame annotations---including lane-validated crossing labels, behavioural states, and COCO-17 format estimated 2D pose keypoints---provides one of the most comprehensive annotation schemas among synthetic pedestrian crossing datasets (Table~\ref{tab:comparison}).

The evolution from PedSynth (947 single-pedestrian clips, stationary ego-vehicle, RGB only) to PedSynth++ (533 multi-pedestrian clips, moving ego-vehicle, three sensor modalities, and 12 weather conditions) demonstrates the framework's extensibility for increasingly complex scenario generation while maintaining full reproducibility through CLI parameterisation and Docker containerisation. 
Future work will add cluster-based parallel generation for scaling, 3D pose lifting from AlphaPose 2D keypoints using LiDAR depth, advanced domain adaptation techniques for sim-to-real transfer, extended multi-agent interaction modelling including pedestrian-cyclist conflicts and social group navigation, and exploration of CARLA's internal skeleton system as an alternative to external pose estimators.


\section*{Declaration of competing interest}
The authors declare that they have no known competing financial interests or personal relationships that could have appeared to influence the work reported in this paper.

\section*{CRediT authorship contribution statement}
\textbf{Muhammad Naveed Riaz:}  Methodology, Software, Investigation, Data curation, Writing – original draft, Visualization.
\textbf{Maciej Wielgosz:} Conceptualization, Methodology, Software, Writing – review \& editing, Supervision.
\textbf{Antonio M.\ L\'{o}pez Pe\~{n}a:} Conceptualization, Funding acquisition, Supervision, Writing – review \& editing.

\section*{Acknowledgements}

\noindent
This project received funding from the European Union's Horizon 2020 research and innovation programme under Marie Sk\l{}odowska-Curie grant agreement No.\ 801342 (Tecniospring INDUSTRY) and the Government of Catalonia's Agency for Business Competitiveness (ACCI\'{O}). We gratefully acknowledge Polish high-performance computing infrastructure PLGrid and the Academic Computer Centre Cyfronet AGH for providing computer facilities and support within computational grant No.\ PLG/2025/018809. This work was supported by the Polish Ministry of Science and Higher Education under subvention funds for the Institute of Electronics AGH.

\section*{Data availability}
The ARCANE-PedSynth framework source code is openly available at \url{https://github.com/wielgosz-info/carla-pedestrians} under the MIT Licence. The PedSynth++ dataset generated using this framework is available from the corresponding author upon reasonable request.

\bibliographystyle{elsarticle-num}
\bibliography{bibliography}

\end{document}